\pdfoutput=1

\documentclass[11pt]{article}

\usepackage[final]{coling}

\usepackage{times}
\usepackage{latexsym}

\usepackage[T1]{fontenc}

\usepackage[utf8]{inputenc}

\usepackage{microtype}

\usepackage{inconsolata}

\usepackage{graphicx}

\usepackage{multirow}

\title{Human vs.\ AI: A Novel Benchmark and a Comparative Study on the Detection of Generated Images and the Impact of Prompts}

\author{Philipp Moe{\ss}ner \\
  Hochschule der Medien \\
  Stuttgart, Germany \\
  \texttt{philippmoessner@gmx.de} \\\And
  Heike Adel \\
  Hochschule der Medien \\
  Stuttgart, Germany \\
  \texttt{adel-vu@hdm-stuttgart.de} \\}

\begin{document}
\maketitle
\begin{abstract}
With the advent of publicly available AI-based text-to-image systems,
the process of creating photorealistic but fully synthetic images has been largely democratized.
This can pose a threat to the public through a simplified spread of disinformation.
Machine detectors and human media expertise can help to differentiate between AI-generated (fake) and real images and counteract this danger.
Although AI generation models are highly prompt-dependent, the impact of the prompt on the fake detection performance has rarely been investigated yet.
This work therefore examines the influence of the prompt's level of detail on the detectability of fake images, both with an AI detector and in a user study.
For this purpose, we create a novel dataset, COCOXGEN, which consists of real photos from the COCO dataset as well as images generated with SDXL and Fooocus using prompts of two standardized lengths. 
Our user study with 200 participants shows that images generated with longer, more detailed prompts are detected significantly more easily than those generated with short prompts.
Similarly, an AI-based detection model achieves better performance on images generated with longer prompts.
However, humans and AI models seem to pay attention to different details, as we show in a heat map analysis.
\end{abstract}
\begin{figure}    \centering\includegraphics[width=1\linewidth]{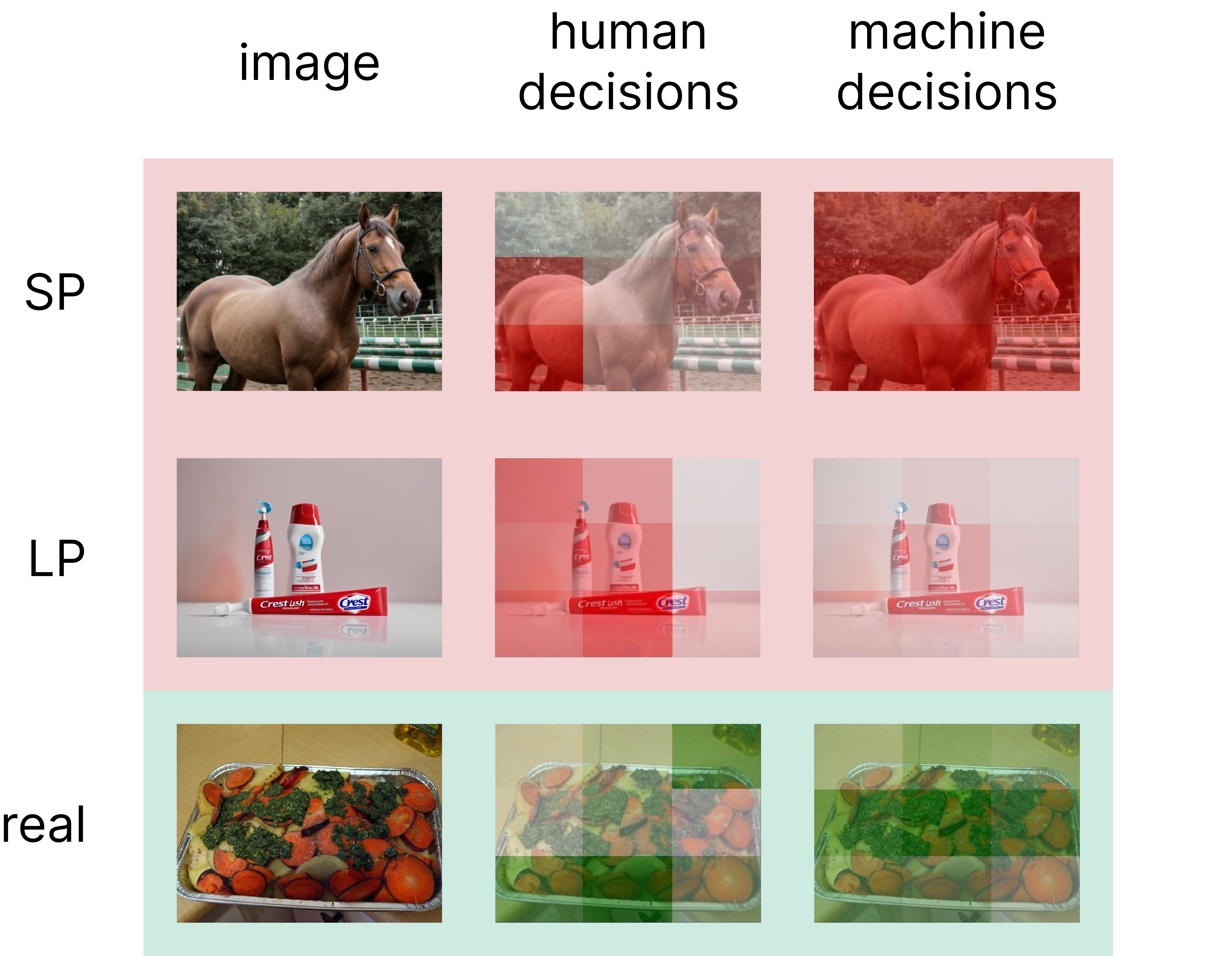}
    \caption{We conduct a study with humans and an AI model detecting real and fake images generated from prompts with a different level of detail (short prompt (SP), long prompt (LP)) and visualize the image areas which led to their decisions.}
    \label{fig:teaser}
\end{figure}
\begin{figure*}[t]
    \centering
    \includegraphics[width=1\linewidth]{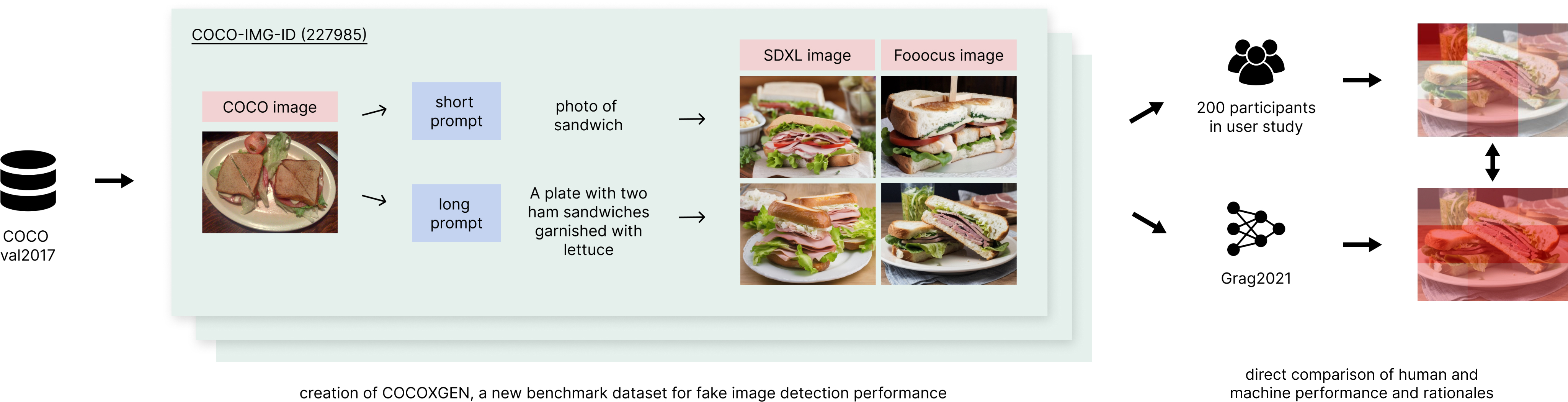}
    \caption{Our contributions are (1) the creation of COCOXGEN, a novel benchmark dataset with images created by two different generation models based on prompts of two different levels of detail, (2) the conduction of a large-scale user study on fake image detection, and (3) a direct comparison of human and machine detection performance and decision rationales.}
    \label{fig:cocoxgen}
\end{figure*}

\section{Introduction}
State-of-the-art AI-based image generators, such as DALL-E \cite{dalle}, Midjourney \cite{midjourney} or Stable Diffusion \cite{stable-diffusion} have the ability to create photorealistic images in a fully synthetic manner \citep{image-perceptual-score}. The fact that these systems are publicly available makes them contribute to the fast spread of synthetic image content on the Internet. This increases the threat of disinformation \citep{disinformation-thorugh-ai}. Thus, there is a need for reliable detection of AI-generated images.

Although it is well known that
image generation is highly dependent on the textual prompt \citep{prompt-design}, 
the impact of the prompt on the fake detection performance has been rarely investigated yet. An exception is the work by \citet{sha2023fake} who analyze prompts regarding their topic and structure.
In particular, the level of detail in the prompt might affect the number of artifacts as more details might force the model to generate an image that deviates more from its training data. Therefore, we pose the question whether the level of detail in the prompt has an impact on the ability of humans and AI-models to detect generated images. 

Existing research on fake image detection shows
that average human performance is not substantially better than chance \citep{cooke_et_al,lu_et_al}. Other works show that machine detection performance can be considered rather reliable, at least for images generated by models that have been included in the training data of detection systems \citep{baraheem_et_al,onlinedetection,corvi_et_al}. While these individual evaluations of humans and AI models show interesting results, it is not possible yet to directly compare human and AI-model performance due to different test setups (i.e. test images).
Thus, another goal of our study is to investigate whether human or AI-model performance dominates.

Some works also investigate which aspects of the input lead to the detection decisions: AI models, on the one hand, tend to involve larger image areas in their decision on real images than on synthetic ones \citep{bird_and_lotfi}. This leads to the impression that the detection of synthetic images is focused on fine details while the detection of real images is focused on more abstract contents. Humans, on the other hand, tend to pay attention to specific objects as well as to their general impression of the image \citep{pocol2023seeing}. Again, it is not possible to directly compare the strategies of humans and AI-models due to different evaluation setups. Thus, we also address the open research question whether humans and AI models consider the same objects and structures in an image when evaluating it as real or fake.

To address our research questions, we present COCOXGEN, a new dataset containing real photos from the COCO dataset \citep{coco} and AI-generated images from SDXL \citep{sdxl} and Fooocus \citep{fooocus} with prompts of two different levels of detail.
Our dataset is publicly available\footnote{\url{https://github.com/heikeadel/cocoxgen}} and can be used in future work as a benchmark dataset for evaluating fake image detection performance.
We conduct a user study with 200 human participants and evaluate the machine-learning classifier Grag2021 \citep{corvi_et_al} on our new dataset to be able to
directly compare the detection performance (F1 scores) of humans and a state-of-the-art AI model.
In our analysis, we visualize image areas that lead to the decisions of humans and machine detectors in a comparable heat map structure
to investigate both qualitatively and quantitatively whether humans and AI models pursue similar strategies.

\section{Related Work}
\subsection{Human Detection Performance}
Previous work investigated human performance on distinguishing AI-generated and real media content by showing humans around 50 real and 50 AI-generated images \citep{cooke_et_al,lu_et_al}. They found an average accuracy of 49\% \citep{cooke_et_al} to 61\% \citep{lu_et_al}. None of the works found a statistically significant effect of previous experience with AI-generated content of the test participants on their accuracy. Still, \citet{lu_et_al} showed a slightly higher performance of participants with previous experience. \citet{pocol2023seeing} additionally investigated how humans come to a classification decision for deepfakes by providing a free text field for explanations. They found that mainly clear suspicious objects and the general impression of the image lead to the decisions. The scale of our user study is comparable to previous works. However, we not only comprehensively evaluate human detection performance including the effect of previous experience with AI-generated content
but also explicitly analyze which image parts lead to the decision of our participants.

\subsection{Machine Detection Performance}
Previous research evaluated different machine-learning approaches to detect AI-generated images \citep{baraheem_et_al,park_et_al,corvi_et_al}. They found that it is possible to achieve high performance in certain conditions \citep{baraheem_et_al,corvi_et_al} but that true generalization to images outside of the scope of the training data remains difficult \citep{onlinedetection}. In addition, downsampling or compressing images decreases detection performance \citep{zhu_et_al}. \citet{bird_and_lotfi} found that the actual objects of the images are of minor importance for the decision of machine detectors.
In contrast to these works, we aim to directly compare the performance of an AI model as well as the image areas that are most relevant for its decision to human performance and decision rationales.

\subsection{Detection Performance Robustness}
Prior work showed that post-generation changes of AI-generated images could considerably decrease the performance of machine-learning detectors \citep{wesselkamp_et_al,carlini_and_farid}.
\citet{wesselkamp_et_al}, for instance, described different approaches of subtracting specific frequencies from the images and \citet{carlini_and_farid} trained a model to calculate optimal perturbations. 
In practical applications, a user would most probably mainly concentrate on altering the content of the generated image via modifications of the prompt. Therefore, we argue that detection models should also be robust against changes in the prompt. The impact of prompts on detection performance has only rarely been  investigated in previous work. \citet{sha2023fake} found that specific words and prompt lengths can lead to lower detection performance. However, they did not investigate these effects in detail. In addition, no prior work has considered human performance when altering AI-generated images. In this paper, we address this research gap and set the impact of the prompt's level of detail as our main research focus.

\section{Dataset}
To the best of our knowledge, there is no dataset publicly available that contains real images and AI-generated images from prompts with a controllable level of detail.
Therefore, we create and publish COCOXGEN (COCO Extended With Generated Images), a novel benchmark dataset for the evaluation of fake image detection performance.
\subsection{Dataset Creation}
We choose the COCO dataset \citep{coco} as the basis for our new dataset because it provides different levels of annotations for the photographs:
several 1-word ``thing'' (objects with a well-defined shape) and ``stuff'' (amorphous background regions) classes as well as 5 complete sentences (captions).
We use those different annotations to build prompts of two different levels of detail as shown in Table \ref{tab:cocoxgen-prompts}. In the following, we refer to the prompt with less detail as ``short prompt (SP)'' and to the prompt with more details as ``long prompt (LP)''. The short prompt is created by prepending ``photo of'' to the most frequent element of the annotated thing and stuff classes. The long prompt is created by selecting the caption with the smallest difference in length to the average length of all captions (10 words). Figure \ref{fig:cocoxgen} shows an examplary COCO image and the two created prompts.

\begin{table}[h]
    \centering
    \begin{tabular}{|c|c|c|}
        \hline
             & \textbf{SP}         & \textbf{LP}           \\ \hline
        \textbf{length} & 3 words    & \(\sim\) 10 words     \\ \hline
        \textbf{shape}  & ``photo of [X]'' & entire sentence  \\ \hline
    \end{tabular}
    \caption{Prompt types used for image generation; both extracted from COCO, the 'X' of the short prompt (SP) is a COCO thing or stuff class (e.g., ``sandwich''), the long prompt (LP) is a COCO caption.}
    \label{tab:cocoxgen-prompts}
\end{table}

For creating the AI-generated images, we use two state-of-the-art methods: SDXL \citep{sdxl}, the latest version of the open source text-to-image model Stable Diffusion \citep{stable-diffusion} and Fooocus \citep{fooocus}, the open-source equivalent to Midjourney \citep{midjourney}.
For Fooocus, we choose the standard model Juggernaut XL V8\footnote{\url{https://huggingface.co/RunDiffusion/Juggernaut-XL-v8}} as the base model and the most popular model for photorealism on CIVITAI,
Realistic Vision V6\footnote{\url{https://civitai.com/models/4201/realistic-vision-v60-b1}}, as the refiner since we aim to generate photorealistic images.
With both generators, we create an image for each prompt. As a result, COCOXGEN's data contains groups of images consisting of 1 real COCO image, its corresponding short and long prompt and 4 generated images (one per prompt type and generation model). 

\subsection{Datasplit and Statistics}
Note that we only use COCO's validation set val2017 as the basis for COCOXGEN to avoid including images which might have been used to train detection models in our benchmark dataset. 
From COCO's validation set, we further remove all images that do not have a 640x480 px resolution 
(standard size of photos in COCO) to ensure that all images have the same size. This is important to be able to exclude the image size as a confounding variable in our experiments. 
While Fooocus is able to generate 640x480 px images natively, SDXL only supports specific resolutions for best image quality. We select 1152x864 px as it results in the same aspect ratio.

Table \ref{tab:cocoxgen-imgs} provides statistics of COCOXGEN. 

\begin{table}[h]
    \centering
    \begin{tabular}{|c|c|c|c|c|}
        \hline
        \textbf{real} & \multicolumn{4}{|c|}{\textbf{AI-generated}} \\
        \cline{1-5}
        1061 & \multicolumn{4}{|c|}{4244} \\
        \hline
        & \multicolumn{2}{|c|}{\textbf{LP}} & \multicolumn{2}{|c|}{\textbf{SP}} \\
        \cline{2-5}
        & \multicolumn{2}{|c|}{2122} & \multicolumn{2}{|c|}{2122} \\
        \cline{2-5}
        & \textbf{Fooocus} & \textbf{SDXL} & \textbf{Fooocus} & \textbf{SDXL} \\
        \cline{2-5}
        & 1061 & 1061 & 1061 & 1061 \\
        \hline
    \end{tabular}
    \caption{Number of images in COCOXGEN (LP: long-prompted images, SP: short-prompted images).}
    \label{tab:cocoxgen-imgs}
\end{table}

\section{User Study}
With this study, we measure human classification performance (in terms of F1 score) for AI-generated and real images. Moreover, we investigate the impact of the level of detail in the prompts on the human performance. 
We assume that a more detailed prompt, i.e., a more complex demand, leads to a higher chance of artifacts in the generated images as the generation model needs to deviate more from its training data to fulfill the individual request. Therefore we state the following hypothesis:
\begin{flushleft}
    $H_1$: \textit{Humans achieve higher detection performance for images generated with longer, more detailed prompts, than for those generated with short prompts.}
\end{flushleft}
Our datasets, which was created using two different state-of-the-art image generation models, further allows us to test human performance per generation model, i.e., to investigate which model creates the most photorealistic images from human perspective.
As we assume that Fooocus images are more photorealistic due to its task-specific refiner, we state the following hypothesis:
\begin{flushleft}
    $H_2$: \textit{Humans achieve higher detection performance for SDXL images than for Fooocus images.}
\end{flushleft}
In addition to investigating pure performance, we further analyze which part of the image leads to the decision of the participants when classifying a photograph as real or fake as well as how certain they are in doing so.
Finally, we analyze whether the participants' experience with AI-generated images before the study influences their detection performance.

\subsection{Study Design}
We recruit 200 participants (127 female, 70 male, 3 non-binary) of ages 14 to 87 years (average age: 25.7). We reached most of them in a university context. As a result, 94.5\% hold at least a high school diploma as their highest educational achievement, and 83.0\% of the participants see AI-generated images sometimes or regularly in their daily lives. 59.5\% have never or just once used image generators themselves. For our study, we randomly select 120 images from COCOXGEN and split them into two disjoint sets of 60 images each (20 real photos, 20 images generated with short prompts, 20 images generated with long prompts, whereby half of the generated images are from SDXL and the other half from Fooocus). Each participants sees one of the sets in the study with the images in random order.
This enables us to test a larger number of images while minimizing possible fatigue effects during the classification process at the same time. 
To ensure diverse content in the images which are used for the user study, we make sure the short prompts (that were created based on the ``stuff'' and ``thing'' classes of the COCO dataset, c.f., Table \ref{tab:cocoxgen-prompts}) do not overlap.
To make sure all images of the study dataset have the same size we further downsample all images generated with SDXL to the size of the COCO and Fooocus images (640x480 px, see above).
For each image, the participants answer the following questions (the actual questionnaire and answer possibilities are provided in Figure \ref{fig:questionnaire} in the appendix):
\begin{enumerate}
  \setlength{\itemsep}{0pt}
  \setlength{\parskip}{0pt}
    \item Is this image real or AI-generated?
    \item How certain are you?
    \item Is there a specific image area which has influenced your decision?
    \item If yes: Which image areas have influenced your decision? (Participants are shown a 3x3 grid on top of the image and are asked to select all fields with decision influence.)
\end{enumerate}

\subsection{Results}
\begin{table}[h]
    \centering
    {\footnotesize
    \begin{tabular}{|c|c|c|c|c|}
    \hline
     \textbf{Subset} & \textbf{Positives} & \textbf{F1} & \textbf{Recall} & \textbf{Precision} \\ \hline 
    All & Real & \textit{$0.7793$} & \textit{$0.8958$} & \textit{$0.6997$} \\ \hline
    All & AI & \textit{$0.8583$} & \textit{$0.7954$} & \textit{$0.9418$} \\ 
    \noalign{\global\arrayrulewidth=1.5pt}
     \hline
     \noalign{\global\arrayrulewidth=0.4pt}
    SP & AI & \textit{$0.8002$} & \textit{$0.7400$} & \textit{$0.8913$} \\ \hline
    LP & AI & \textit{$0.8697$} & \textit{$0.8508$} & \textit{$0.9006$} \\
     \noalign{\global\arrayrulewidth=1.5pt}
     \hline
     \noalign{\global\arrayrulewidth=0.4pt}
    Fooocus & AI & \textit{$0.7857$} & \textit{$0.7190$} & \textit{$0.8880$} \\ \hline
    SDXL & AI & \textit{$0.8822$} & \textit{$0.8718$} & \textit{$0.9030$} \\ \hline
    COCO & Real & \textit{$0.7793$} & \textit{$0.8958$} & \textit{$0.6997$} \\ \hline
    \end{tabular}}
    \caption{Average F1, recall and precision scores of all study participants for specific subsets of the study images; ``Positives'' indicates the class for which the scores were calculated.}
    \label{table:study-eval-overview}
\end{table}

\begin{figure}[h]
    \includegraphics[width=\linewidth]{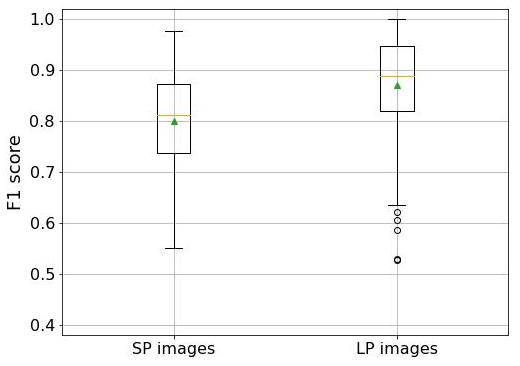}
    \caption{F1 scores per prompt type (SP: short prompt with less details, LP: long prompt with more details).}
    \label{fig:f1-by-prompt-length-study}
\end{figure}

\paragraph{Impact of prompt on detection performance.}
Figure \ref{fig:f1-by-prompt-length-study} shows that participants have higher detection performance for images generated with long prompts (LP) than for images generated with short prompts (SP). In particular, the average and median performance of the participants are higher for LP images than for SP images.
Thus, the participants recognized LP images more easily as being AI-generated than SP images.
A one-sided Wilcoxon test on our data shows that this difference is statistically significant ($p = 2.1696^{-22}$),
indicating that our hypothesis $H_1$ is valid. Furthermore, the difference in F1 scores between the image groups of different prompt lengths can be categorized as strong with an effect size of Cohen's $d = 0.8809$.

\begin{figure}
    \centering
    \includegraphics[width=1\linewidth]{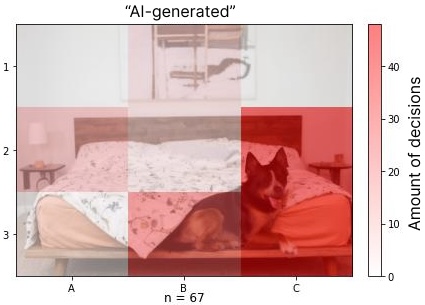}
    \caption{The image area selection results show that participants pay particular attention to clear objects.}
    \label{fig:study-heatmap-fake-example}
\end{figure}

\paragraph{Analysis of relevant image areas.}
Overall, slightly more of the participants' decisions were based on the ``general impression'' (48.78\%) rather than on specific image areas (45.16\%).\footnote{For the remaining 6.05\% of images, participants noted they were unsure.}
For real images, more decisions were based on the ``general impression'' (for 65.23\% of all real images and 71.45\% of correctly classified real images).
For AI-generated images, concrete image areas slightly outweigh the general impression for decision-making (for 54.40\% of all AI-generated images and 65.28\% of correctly classified AI-generated images participants named specific image areas).
This indicates that the participants were able to detect suspicious areas or concrete artifacts rather in AI-generated images than in real images. 
Interestingly, for 63.31\% of real images that were incorrectly classified as AI-generated, participants named concrete image areas as decision rationales. This shows that participants look for suspicious areas in the image in order to classify an image as ``AI-generated'' while the decision to classify an image as ``real'' rather depends on the general impression.
By directly analyzing the individual heat maps for each image, we observe that the study participants pay particular attention to objects that can be clearly separated in the image and use these as the basis for their decision (instead of, e.g., background structures). In case of AI-generated images, obvious artifacts are selected as well (see Figure \ref{fig:study-heatmap-fake-example} for an example).
This finding is in line with previous work which showed that people generally concentrate on structures in a picture that strongly stand out \citep{parkhurst2002modeling}.

\begin{figure}
    \centering
    \includegraphics[width=.97\linewidth]{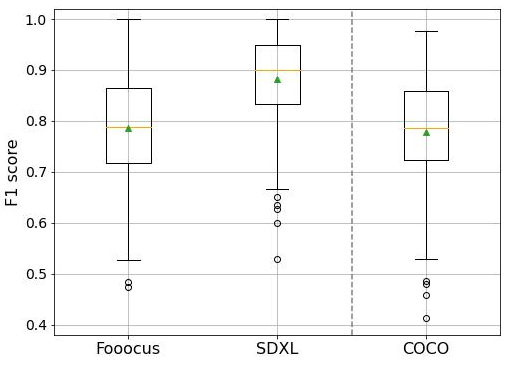}
    \caption{F1 scores per image generator (left and middle; the lower, the more realistic are the generated images) and for real COCO images (right; the higher the better)}
    \label{fig:f1-by-image-source-study}
\end{figure}
\paragraph{Impact of generation model on detection performance.}
The lower the F1 score for detecting AI-generated images, the more photorealistic the images are.
When looking at the generator-specific F1 scores in Figure \ref{fig:f1-by-image-source-study}, we observe lower F1 scores for Fooocus images than for SDXL images. 
The study participants therefore tended to recognize the SDXL images more easily  as AI-generated than the Fooocus images, which indicates that Fooocus produces more photorealistic results than SDXL. 
This result is also statistically significant (one-sided Wilcoxon test, $p = 5.1944e^{-29}$).
This indicates that our hypothesis $H_2$ is valid. Furthermore, the difference in F1 scores between different generators can be categorized as strong with an effect size of Cohen's $d = 1.1978$.
Interestingly, the F1 scores of assigning COCO photos to their correct class (i.e., ``real'') are in a similar range as the F1 scores of assigning Fooocus images to their correct class (i.e., ``AI''). This is a result of a lower precision for the ``real'' class, i.e., many AI-generated images were falsely classified as real. Considering only recall, COCO photos were best assigned to the correct class (see Table \ref{table:study-eval-overview}).

\begin{figure}
    \centering
    \includegraphics[width=\linewidth]{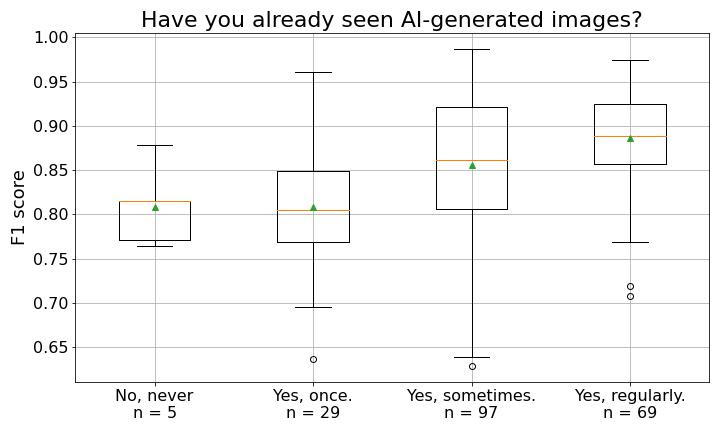}
    \caption{F1 scores for AI-generated images by experience.}
    \label{fig:f1-by-watch-experience}
\end{figure}
\begin{figure}
    \centering
    \includegraphics[width=\linewidth]{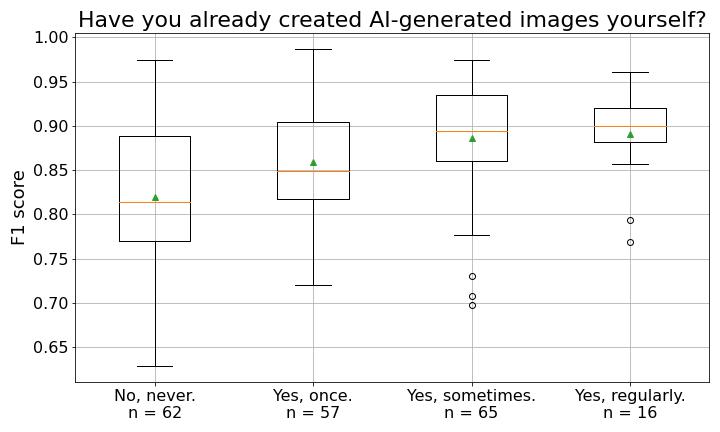}
    \caption{F1 scores for AI-generated images by  experience with image generation.}
    \label{fig:f1-by-generation-experience}
\end{figure}

\paragraph{Impact of experience on detection performance.}
Participants who rated their experience with AI-generated images higher also tended to perform better at detecting AI-generated images (Figures \ref{fig:f1-by-watch-experience} and \ref{fig:f1-by-generation-experience}).
Interestingly, the boxes (interquartile ranges) for ``no, never'' and ``yes, once'' as well as the boxes for ``yes, sometimes'' and ``yes, regularly'' fully overlap while there is a clearer difference between ``yes, once'' and ``yes, sometimes''. 

\begin{figure}
    \centering
    \includegraphics[width=.85\linewidth]{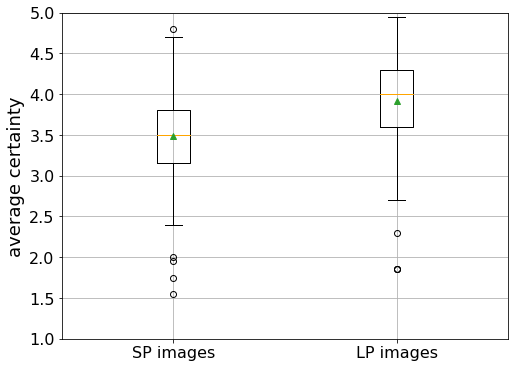}
    \caption{Average decision certainty of the participants for images generated by long prompts with more details (LP) vs.\ short prompts (SP).}
    \label{fig:certainty-by-prompt-length}
\end{figure}

\paragraph{Analysis of participants' decision certainty.}
We cannot find a considerable difference in the decision certainty (values ranging from 1 (``very uncertain'') to 5 (``very certain'')) of the participants for AI-generated (average of 3.70) and real images (average of 3.80).
When comparing their average decision certainty for LP  images (average of 3.91) and SP images (average of 3.49) (c.f., Figure \ref{fig:certainty-by-prompt-length}), the difference is larger. A one-sided Wilcoxon test on this data shows that this difference is also statistically significant ($p = 1.2486e^{-29}$). 
Thus, the participants not only performed better at classifying images generated from more detailed prompts but were also more certain in their decisions 
compared to images generated from short prompts.

\section{Machine Classification Performance}
For automatically distinguishing real from AI-generated images, we apply
Grag2021 \citep{gragnaniello2021gan}, a ResNet50 \citep{resnet} which is one of the most popular architectures for image classification.
We use the trained and publicly available version by \citet{corvi_et_al}.
Grag2021 has been trained on COCO and Latent Diffusion images which fits our dataset well.

\subsection{Experimental Setup}
We test Grag2021 on all 120 images which have also been selected for the study with the human participants. Grag2021 returns a feature map of 60x80 logit values whereby negative values indicate class ``real'' and positive values indicate class ``AI-generated''. We follow \citet{corvi_et_al} and obtain a single output value by averaging these logits. To transfer the output to a probability value, we apply the sigmoid function.
The resulting value $x$ indicates the probability for class ``AI-generated''. Therefore, if $x \geq 0.5$, the classification result is ``AI-generated'', if $x < 0.5$, the resulting class is ``real''.\footnote{Note that we used the default threshold of 0.5 from logistic regression for the classification decision.}
For evaluation, we calculate the F1 scores of the detector
and visualize its output feature map as a decision heat map for each image.\footnote{To ensure that the heat map has the same resolution as the input image, we let each value of the output feature map correspond to an area of 8x8 pixels in the heat map.} To compare the model's heat maps with the study participants' image area selections, we apply the same 3x3 grid of our study on the detector feature maps. For each grid field, we add all positive and all negative values separately, resulting in two distinct heat maps: one indicating rationales for class ``real'' (per-field sum of negative values) and one indicating rationales for class ``AI-generated'' (per-field sum of positive values).

\subsection{Results}
\begin{table}
    \centering
    {\footnotesize
    \begin{tabular}{|c|c|c|c|c|}
    \hline
    \textbf{Subset} & \textbf{Positives} & \textbf{F1} & \textbf{Recall} & \textbf{Precision} \\ 
    \noalign{\global\arrayrulewidth=1.5pt}
    \hline
    \noalign{\global\arrayrulewidth=0.4pt}
    All & Real & \textit{$0.6957$} & \textit{$1.0000$} & \textit{$0.5333$} \\ \hline
    All & AI & \textit{$0.7200$} & \textit{$0.5625$} & \textit{$1.0000$} \\
    \noalign{\global\arrayrulewidth=1.5pt}
     \hline
     \noalign{\global\arrayrulewidth=0.4pt}
    SP & AI & \textit{$0.7097$} & \textit{$0.5500$} & \textit{$1.0000$} \\ \hline
    LP & AI & \textit{$0.7302$} & \textit{$0.5750$} & \textit{$1.0000$} \\
    \noalign{\global\arrayrulewidth=1.5pt}
     \hline
     \noalign{\global\arrayrulewidth=0.4pt}
    Fooocus & AI & \textit{$1.0000$} & \textit{$1.0000$} & \textit{$1.0000$} \\ \hline
    SDXL & AI & \textit{$0.2222$} & \textit{$0.1250$} & \textit{$1.0000$} \\ \hline
    COCO & Real & \textit{$0.6957$} & \textit{$1.0000$} & \textit{$0.5333$} \\ \hline
\end{tabular}}
    \caption{F1, recall and precision scores of Grag2021 for specific subsets of the study images. ``Positives'' indicates the class for which the scores were calculated.}
    \label{table:detector-test-eval-overview}
\end{table}

\begin{figure*}[h]
    \centering
    \includegraphics[width=.9\linewidth]{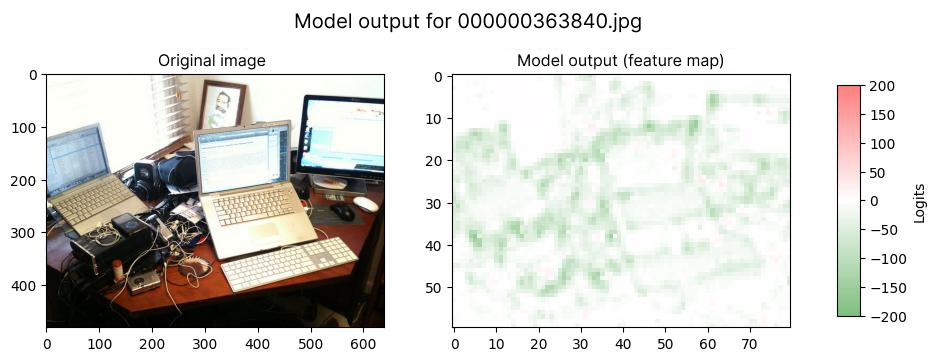}
    \caption{Visualization of the model's output feature map of 60x80 logit values for a real image from COCOXGEN.}
    \label{fig:featuremap-real}
\end{figure*}

Table \ref{table:detector-test-eval-overview} shows the results of the AI detection model Grag2021. It performs 
16.11\% worse in detecting AI-generated images than the average human participant.
As described before, the images generated by SDXL were downsampled for the user study. We, therefore, compare the performance on the study images and on the original images (Table \ref{table:detector-test-study-vs-original} in appendix). While the performance on original images is very high (F1 score of 0.9744), the performance on the study images is considerably lower (0.7222) which can be explained with a very poor performance on downsampled SDXL images (0.2222). This is in line with the work of \citet{zhu_et_al} who found that downsampling or compressing images decreases detection performance.

\paragraph{Impact of prompt on detection performance.}
Similar to the results of our user study, the detection performance on images generated with long and detailed prompts (LP) is higher than the performance on images generated with short prompts (SP). Note that this difference is independent of downsampling but not statistically significant (according to a permutation test). 

\paragraph{Analysis of relevant image areas.}
Many heat maps for real images (around 57.50\%) show clearly recognizable object structures or edges (Figure \ref{fig:featuremap-real}). This means that the object edges have particularly high activation values compared to the rest of the image.
For the remaining images including most of the AI-generated images (70.00\%), the heat maps depict a relatively uniform noise (an exemplary heat map is provided in the appendix, Figure \ref{fig:featuremap-ai-image}).

When comparing the detector's heat maps with those from the study participants, we observe overlaps for some images (c.f., Figure \ref{fig:activation-selection-comparison} in appendix).
To quantify this overlap, we calculate two Spearman correlation coefficients for each image (one for overlaps of areas indicating class ``AI-generated'' and one for overlaps of areas indicating class ``real'').
The correlation coefficient for class ``AI-generated'' corresponds to $\rho =$ 0.1988 (p-value: 0.3921), the correlation coefficient for class ``real'' is  $\rho =$ 0.2287 (p-value: 0.4269). To conclude, a low but non-significant correlation was found between the image areas that were most relevant for the participants' and detector's decisions.

\paragraph{Impact of generation model on detection performance.}
Table \ref{table:detector-test-eval-overview} shows that real images and fake images generated by Fooocus are recognized perfectly by the detector when considering recall. Considering F1 scores, Fooocus images remain perfectly distinguished from the others while 
AI-generated images with SDXL are sometimes mistakenly categorized as ``real'', influencing the recall on the SDXL subset as well as the precision for the COCO class.

\section{Discussion}
This section discusses the implications of the most important findings of this work.

\subsection{Perspective for Real-World Applications}
Overall, the machine detector performed worse in distinguishing real from AI-generated study images than the average human study participant which can be attributed to partially downsampling the AI-generated images.
The almost perfect results of the machine detector under optimal conditions (no downsampling) indicates a clear opportunity for machine detectors to recognize AI-generated images even when they appear photorealistic to humans.
Unfortunately, downsampling is necessarily carried out on many internet platforms, weakening the practical applicability of current machine detectors.

\subsection{Level of Detail in Prompt}
Our study shows that the detection performance for images generated with long and detailed prompts is significantly higher than for images generated with short prompts. We hypothesize that the image generator has to deviate more from its training data to fulfil the needs of a complex prompt with many details. This might lead to more artifacts in the generated image.
Thus, not only the technical implementation influences the quality and detectability of the output images but also the prompt that was used for generation.

\subsection{Impact of Experience}
Study participants who stated that they were more experienced in viewing and creating AI-generated images also tended to be better at detecting them.
This suggests that the ability to recognize AI-generated images and distinguish them from real photos can be trained.

\subsection{Rationales for Decisions}
The percentage of instances for which participants declared the ``general impression'' as the main decision argument shows that visible artifacts do not need to be present in the image for people to be skeptical about the authenticity of an image.
This can be seen as an opportunity for the human ability to detect AI-generated images, especially in light of the fact that generation algorithms are constantly evolving and will produce fewer and fewer visible artifacts in the future. 
While humans mainly concentrate on clearly distinguishable objects in the images \citep{parkhurst2002modeling}, 
the machine detector gives high activations mainly to fine details (high frequencies) or object edges.
This difference might also be the reason why we could not find a significant overlap between the selected image areas of humans and the most activated ones by the machine detector. 
An interesting direction could be to combine the complementary attention areas of humans and machines in a collaborative setting.

\subsection{Limitations}
Findings of this work cannot be generalized to all image generation models, especially since both generators of this work are derived from Latent Diffusion without considering GAN methods or transformer architectures. 
Similarly, only a single machine detection model was tested.
Since we performed a user study, we also used only 120 images from COCOXGEN in this work, while the entire dataset consists of 5305 images. 

Although we observed statistically significant correlations in our user study, it is important to note that the age, gender and education distribution of our participants does not reflect the distribution of the whole population. The same holds for the prior experience with image generation which was rather high for our participants on average as most of them were recruited in a university context.

\section{Conclusion and Future Work}
In this work, we investigated the influence of the prompt's level of detail for distinguishing AI-generated images from real ones.
In particular, we explored both the performance of humans and a machine detector and directly compared their decision rationales. For this purpose, we created a novel benchmark dataset COCOXGEN, which contains AI-generated SDXL and Fooocus images created with prompts of two different levels of detail as well as real COCO images. We found that images generated with prompts with more details can be recognized more easily as fake than images generated with short prompts with less details. 
This observation holds for both humans and the machine detector although their decision rationales show only a low correlation.

Future work can expand our study by investigating more types of generation and machine detection models. In addition, additional aspects of prompt composition (e.g., the number of words per part-of-speech class) and their impact on fake image detection performance can be explored.

\section*{Acknowledgments}
We thank the Etzold foundation for supporting the publication of this work.
We thank all participants of our study for their time and their support of our research.
We further thank all reviewers for their helpful feedback.

\bibliography{coling_latex}

\appendix
\pagebreak
\onecolumn 
\section{Appendix}
\label{sec:appendix}

\subsection{Questionnaire}
\begin{figure*}[h]
    \centering
    \includegraphics[width=.87\linewidth]{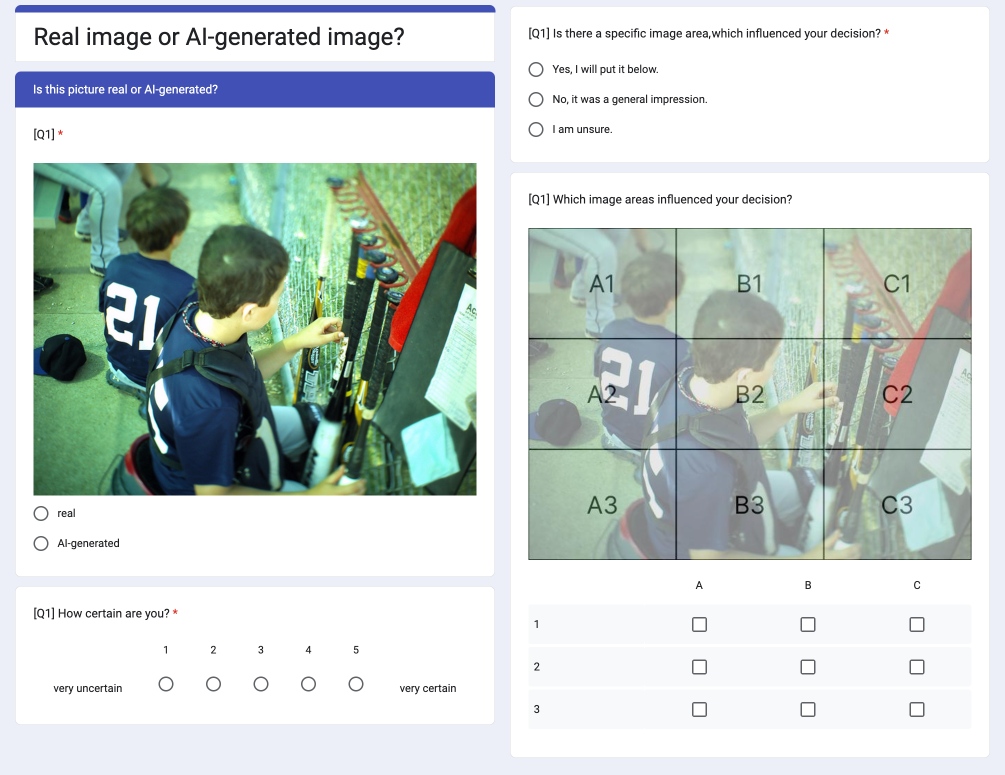}
    \caption{Questions and answer possibilities for image classification in our user study.}
    \label{fig:questionnaire}
\end{figure*}

Figure \ref{fig:questionnaire} shows the repeated structure of questions for each image presented to the study participants.

\subsection{Comparing model results on images with and without downsampling}
Table \ref{table:detector-test-study-vs-original} shows the difference in the model's detection performance for downsampled (study images) and original images. Note that it is the downsampled SDXL images which are influencing the overall worse performance of the model on the study dataset. The low recall for the downsampled SDXL images (0.1250) indicates that a lot of these images were incorrectly classified as real. This influences the precision for real images weakening the corresponding F-score.
\begin{table}[h]
    \centering
    \begin{tabular}{|c|c|c|c|c|c|c|c|}
    \hline
    \multirow{2}{*}{\textbf{Subset}} & \multirow{2}{*}{\textbf{Positives}} & \multicolumn{3}{c|}{\textbf{Study images}} & \multicolumn{3}{c|}{\textbf{Original images}} \\ 
    \cline{3-8}
    & & \textbf{F1} & \textbf{Recall} & \textbf{Precision} & \textbf{F1} & \textbf{Recall} & \textbf{Precision} \\ 
    \noalign{\global\arrayrulewidth=1.5pt}
    \hline
    \noalign{\global\arrayrulewidth=0.4pt}
    All & Real & \textit{$0.6957$} & \textit{$1.0000$} & \textit{$0.5333$} & \textit{$0.9524$} & \textit{$1.0000$} & \textit{$0.9091$} \\ \hline
    All & AI & \textit{$0.7200$} & \textit{$0.56250$} & \textit{$1.0000$} & \textit{$0.9744$} & \textit{$0.9500$} & \textit{$1.0000$} \\
    \noalign{\global\arrayrulewidth=1.5pt}
    \hline
    \noalign{\global\arrayrulewidth=0.4pt}
    SP & AI & \textit{$0.7097$} & \textit{$0.5500$} & \textit{$1.0000$} & \textit{$0.9610$} & \textit{$0.9250$} & \textit{$1.0000$} \\ \hline
    LP & AI & \textit{$0.7302$} & \textit{$0.5750$} & \textit{$1.0000$} & \textit{$0.9873$} & \textit{$0.9750$} & \textit{$1.0000$} \\
    \noalign{\global\arrayrulewidth=1.5pt}
    \hline
    \noalign{\global\arrayrulewidth=0.4pt}
    Fooocus & AI & \textit{$1.0000$} & \textit{$1.0000$} & \textit{$1.0000$} & \textit{$1.0000$} & \textit{$1.0000$} & \textit{$1.0000$} \\ \hline
    SDXL & AI & \textit{$0.2222$} & \textit{$0.1250$} & \textit{$1.0000$} & \textit{$0.9474$} & \textit{$0.9000$} & \textit{$1.0000$} \\ \hline
    COCO & Real & \textit{$0.6957$} & \textit{$1.0000$} & \textit{$0.5333$} & \textit{$0.9524$} & \textit{$1.0000$} & \textit{$0.9091$} \\ \hline
    \end{tabular}
    \caption{F1, recall and precision scores of Grag2021 for specific subsets of the study images. ``Positives'' indicates the class for which the F1 score was calculated. ``Study images'' refers to the results on the study images, ``Original images'' to the results on the corresponding images without downsampling.}
    \label{table:detector-test-study-vs-original}
\end{table}

\subsection{Model output for an AI-generated image}
In contrast to Figure \ref{fig:featuremap-real}, which visualizes the output of the model for a real image, here we can see the output for an examplary AI-generated image. Note the relatively uniform noise in contrast to the clearly visible object edges in Figure \ref{fig:featuremap-real}.
\begin{figure*}[h]
    \centering
    \includegraphics[width=.83\linewidth]{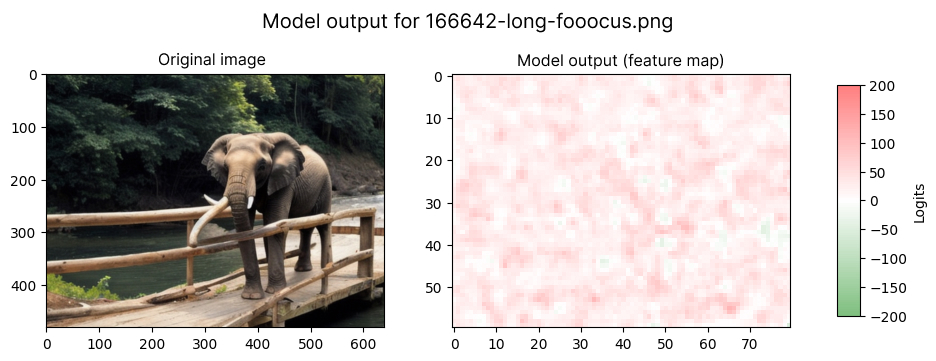}
    \caption{Visualization of the model's output feature map of 60 x 80 logit values for an AI-generated image from the study dataset}
    \label{fig:featuremap-ai-image}
\end{figure*}
\subsection{Image area selection of the model and the participants}
\begin{figure*}[h]
    \centering
    \includegraphics[width=.83\linewidth]{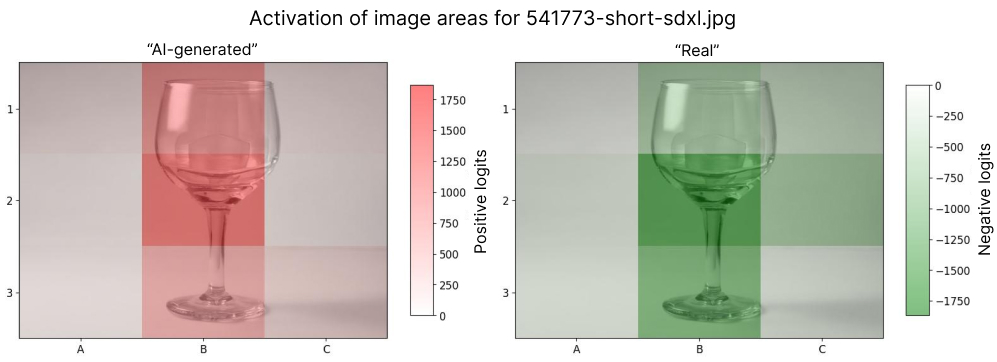} \\
    \vspace{1cm}
    \hspace{-.28cm}
    \includegraphics[width=.83\linewidth]{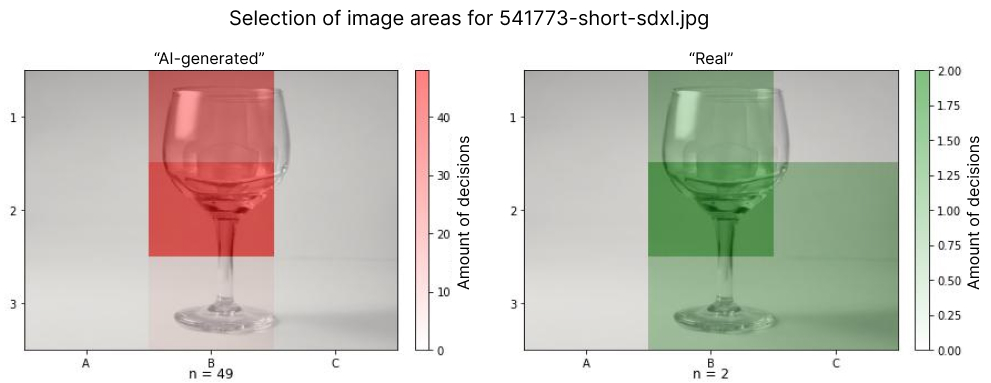}
    \caption{Comparison of the most activated image areas by the detector (each grid field sums up the corresponding logit values from the detector feature map) (top) and the most selected image areas by the test participants (bottom)}
    \label{fig:activation-selection-comparison}
\end{figure*}

Figure \ref{fig:activation-selection-comparison} shows the most selected / activated image areas for an examplary image from the study dataset. Note that we can see an overlap of the most activated image areas by the model and the most selected ones by the study participants.

\end{document}